\documentclass[a4paper]{report}
\usepackage[utf8]{inputenc}
\usepackage[T1]{fontenc}
\usepackage{RJournal}
\usepackage{amsmath,amssymb,array}
\usepackage{booktabs}

\begin{document}

%% do not edit, for illustration only
\sectionhead{Contributed research article}
\volume{XX}
\volnumber{YY}
\year{20ZZ}
\month{AAAA}

%% replace RJtemplate with your article
\begin{article}
  % !TeX root = RJwrapper.tex

\title{\pkg{text2sdg}: An R package to Monitor Sustainable Development Goals from Text}
\author{by Dominik S. Meier, Rui Mata, and Dirk U. Wulff}

\maketitle

\abstract{
Monitoring progress on the United Nations Sustainable Development Goals (SDGs) is important for both academic and non-academic organizations. Existing approaches to monitoring SDGs have focused on specific data types; namely, publications listed in proprietary research databases. We present the \CRANpkg{text2sdg} package for the R language, a user-friendly, open-source package that detects SDGs in any kind of text data using different existing or custom-made query systems. The \CRANpkg{text2sdg} package thereby facilitates the monitoring of SDGs for a wide array of text sources and provides a much-needed basis for validating and improving extant methods to detect SDGs from text.
}

\section{Introduction}

The United Nations Sustainable Development Goals (SDGs) have become an important guideline for both governmental and non-governmental organizations to monitor and plan their contributions to social, economic, and environmental transformations. The 17 SDGs cover large areas of application, from ending poverty and improving health, to fostering economic growth and preserving natural resources. As the latest UN report \cite[]{SGD_report2022} attests, the availability of high-quality data is still lacking in many of these areas and progress is needed in identifying data sources that can help monitor work on these goals. Monitoring of SDGs has typically been based on economic and health data (e.g., \url{https://sdg-tracker.org/}; \url{https://www.sdgindex.org/}), which are often difficult and costly to gather. One attractive alternative that has emerged from recent scientometric efforts is to detect SDGs from text, such as academic publications. Digitized text represents an attractive resource for monitoring SDGs across a large number of domains because it is becoming widely available in various types of documents, such as news articles, websites, corporate reports, and social media posts. In light of this promise, we developed \CRANpkg{text2sdg}, a freely available, open-source tool to enable the SDG-labeling of digitized text and facilitate methodological development in this area. In what follows, we first present some background on existing labeling systems developed to identify SDGs from text, and then provide an overview of the \CRANpkg{text2sdg} package, showcase its use in a representative case study, and discuss the promise and limitations of the approach. 

\section{An overview of SDG labeling systems}

 The \CRANpkg{text2sdg} package provides a user-friendly way to use any existing or custom-made labeling system developed to monitor the 17 SDGs in text sources. In the following, we will briefly introduce five existing labeling systems implemented in \CRANpkg{text2sdg}: the Elsevier, Aurora, SIRIS, OSDG, and SDSN systems. See table \ref{tab:systems_overview} for overview of these labeling systems. We address custom-made labeling systems in a dedicated section below.   
 
 The most prominent SDG labeling systems has been developed by \textit{Elsevier}. The Elsevier labeling system was integrated into the Times Higher Education Impact Rankings in 2019, which at the time compared 1,118 universities in their efforts to address the SDGs as measured by the frequency of SDG-related terms in their academic output. The Elsevier queries consist of a list of expert-vetted keywords that are combined using logical AND operators, implying that multiple keywords must be met to label a document as containing a certain SDG. The development of the queries started with an original list of keywords for each SDG that were iteratively fine tuned to maximize the number of identified papers closely reflecting the different SDGs. This involved cropping or combining keywords to reduce the number of irrelevant hits. A detailed report on the initial development of the Elsevier query system is provided by \citet[]{jayabalasingham2019identifying}. Since the first version, the Elsevier labeling system has been iteratively improved, with the latest versions including additional information specific to academic publications and the Scopus database, such as identifiers of journal names or research areas. \CRANpkg{text2sdg} implements the latest version without such additional identifiers to broaden the package's applicability beyond the Scopus database \citep[]{jayabalasingham2019identifying}.
 
 The Aurora Universities Network's "Societal Impact and Relevance of Research" working group started to develop a labeling system in 2017 to increase the visibility of research into the SDGs. Aurora's queries were developed with the goal of identifying SDG-related academic publications included in the Scopus database. Consequently, the syntax of Aurora queries is similar to the Scopus query language and the Elsevier system. However, in contrast to the Elsevier system, the queries combine  keywords in a more complex fashion, recruiting Boolean- (AND, OR) and proximity operators (e.g., w/3, implying within 3 words). As a result, Aurora's keywords are more specific, possibly leading to a smaller number of false positives. The initial version of the Aurora system only included terms that appear in the SDG policy text of the targets and indicators defined by the United Nations. Subsequent versions expanded on this by including additional keywords that reflect academic terminology. \CRANpkg{text2sdg} implements version 5.0 of the Aurora labeling system \citep{vanderfeesten_maurice_2020_3817445}. This version represents an improvement on previous versions based on a survey study \citep{vanderfeesten_maurice_2020_3813230} and modifications inspired in other efforts, namely those from Elsevier (above) and SIRIS (introduced next). 

 The SIRIS labeling system \cite[]{duran_silva_nicolau_2019_3567769} was created by SIRIS Academic as part of the  \href{http://science4sdgs.sirisacademic.com/}{"science4sdgs"} project to better understand how science, innovation efforts, and technology related to the SDGs. The SIRIS queries were constructed in a five-step procedure. First, an initial list of keywords was extracted from the United Nations official list of goals, targets and indicators. Second, the list was manually enriched on a basis of a review of SDGrelevant literature. Third, a word2vec model that was trained on a text corpus created from the enriched keyword list was used to identify keywords that were semantically related to the initial list. Fourth, using the DBpedia API, keywords were added that, according to the Wikipedia corpus, had a categorical relationship with the initial list. Fifth, and finally, the keyword list was manually revised. The queries of the SIRIS labeling system primarily consist of individual keywords that occasionally are combined with a logical AND. \CRANpkg{text2sdg} implements the only currently available version of the SIRIS labeling system \cite[]{duran_silva_nicolau_2019_3567769} . 

The Open Source SDG (OSDG) project combines data from multiple sources to detect SDGs in text. Instead of developing yet another query system, OSDG's aim was to re-use and integrate existing knowledge by combining multiple SDG "ontologies" (i.e., query systems). OSDG has also made use of Microsoft Academic Graph to improve their results but because our query-based system cannot implement this procedure, we adopt the simpler ontology initially proposed by OSDG. The labeling system was based on central keywords in the SDG United Nations description (e.g."sanitation" was classified into "SDG6") and then manually expanded with additional relevant keywords identified from a corpus of already labeled documents. The resulting keyword list only makes use of the OR operator. \CRANpkg{text2sdg} implements the only currently available version of these queries \cite[]{Bautista2019}. 

 Finally, the Sustainable Development Solutions Network \cite[SDSN,][]{sdsn} labeling system contains SDG-specific keywords compiled in a collaborative effort by several universities from the Sustainable Development Solutions Network (SDSN) Australia, New Zealand \& Pacific Network. This query system was developed to detect SDGs in large sets of university-related text data, such as course listings or research publications. The authors used United Nations documents, Google searches, and personal communications as sources for the keywords. This query system combines keywords with OR operators and does not make use of AND operators. 

 All in all, as can be seen in Table \ref{tab:systems_overview}, the latter systems differ from the former three in the complexity of their queries: the Elsevier, Aurora, and SIRIS systems make use of keyword-combination queries and other criteria, such as proximity operators, whereas OSDG and SDSN only make use of keywords. As we will see in the results below, this has implications for the number of labels produced by the different systems.

\begin{table}[]
\resizebox{\columnwidth}{!}{%
\begin{tabular}{@{}llllll@{}}
\toprule
Labeling system &
  SDGs covered &
  Query operators &
  
  Unique keywords per &
  Example query (SDG-01) &
  \\ & & &
  SDG (mean \& SD) &
  
   \\ \midrule
Elsevier &
  SDG 1 - SDG 16 &
  OR, AND, wildcards &
  74.9 (21.7) &
  "extreme poverty" &
   \\
Aurora &
  SDG 1 - SDG 17 &
  OR, AND, wildcards, proximity search&
  89.6 (31.6) &
  ("poverty") W/3 ("chronic*" OR "extreme") &
   \\
SIRIS &
  SDG 1 - SDG 16 &
  OR, AND &
  262 (148) &
  ("anti-poverty") AND ("poverty" OR "vulnerability") &
   \\
OSDG &
  SDG 1 - SDG 17 &
  OR &
  245 (236) &
  "absolute poverty" &
   \\
SDSN &
  SDG 1 - SDG 17 &
  OR &
  62.6 (16.6) &
  "End poverty" &
  \\ \bottomrule
\end{tabular}%
}
\caption{Overview of the Labeling Systems Implemented in \CRANpkg{text2sdg}. Legend: OR---keywords are combined using logical ORs, implying that only the keywords must be matched to assign an SDG label; AND---keywords are combined using logical ANDs, implying that multiple keywords must be matched to assign an SDG label; wildcards---keywords are matched considering different keyword parts; proximity search---keywords must co-occur within a certain word window to assign an SDG label.}
\label{tab:systems_overview}
\end{table}

\section{The text2sdg package}

\subsection{Designing text2sdg}

Despite the effort put into developing various labeling systems and their great promise in addressing the SDG-related data scarcity, extant implementations of these approaches are not without shortcomings. First, the labeling systems were mostly developed to be used within academic citation databases (e.g., Scopus) and are not easily applied to other text sources. Second, existing implementations lack transparent ways to communicate which features are matched to which documents or how they compare between a choice of labeling systems. We alleviate these shortcomings by providing an open-source solution, \CRANpkg{text2sdg}, that lets users detect SDGs in any kind of text using any of the above-mentioned systems or, even, customized, user-made labeling systems. The package provides a common framework for implementing the different extant or novel approaches and makes it easy to quantitatively compare and visualize their results. 

\subsection{Overview of \CRANpkg{text2sdg} package}

At the heart of the \CRANpkg{text2sdg} package are the Lucene-style queries that are used to detect SDGs in text. These queries map text features (i.e., words or a combination of words) to SDGs. For example, a text that contains the words "fisheries" and "marine" would be mapped to SDG 14 (i.e., conserve and sustainably use the oceans, seas and marine resources for sustainable development) by the Aurora system. To enable the use of such queries in R, the \CRANpkg{text2sdg} package recruits the \CRANpkg{corpustools} \citep{corpustools}. \CRANpkg{corpustools} has been built to implement complex search queries and execute them efficiently for large amounts of text. Based on this, \CRANpkg{text2sdg} provides several functions that implement extant labeling systems, facilitate the specification of new labeling systems, and analyze and visualize search results. Table \ref{tab:functions_overview} gives an overview of the \CRANpkg{text2sdg} core functions. 

The main function of \CRANpkg{text2sdg} is \code{detect\_sdg}. The function runs some or all of the implemented labeling systems to identify SDGs in texts. The texts are provided to the function via the \code{text} argument as either a character vector or an object of class \code{"tCorpus"} from \CRANpkg{corpustools}. All other arguments are optional. By default, the function runs only the Aurora, Elsevier, and SIRIS labeling systems to identify all 17 SDGs. However, both the the systems and the SDGs can be customized using the \code{system} and \code{sdgs} arguments, respectively. The function returns a \code{tibble} that includes one row per hit and has the following columns (and types):

\begin{itemize}
  \item document (factor) - index of element in the character vector or corpus supply for text
  \item sdg (character) - labels indicating the matched SDGs
  \item system (character) - the query system that produced the match
  \item query\_id (integer) - identifier of query in the query system
  \item features (character) - words in the document that were matched by the query
  \item hit (numeric) - running index of matches for each query system
\end{itemize}

Further details on the \code{detect\_sdg} function and its output will be presented in the next section.

The \code{detect\_any} function implements the same functionality as \code{detect\_sdg}, but permits the user to specify customized or self-defined queries. These queries are specified via the code \texttt{queries} argument and must follow the syntax of the \CRANpkg{corpustools} package. See Practical Considerations section for more details).

To support the interpretation of SDG labels generated by \code{detect\_sdg} and \code{detect\_any}, \CRANpkg{text2sdg} further provides the \code{plot\_sdg} and \code{crosstab\_sdg}  functions. The \code{plot\_sdg} function visualizes the distribution of SDG labels identified in documents by means of a customizable barplot showing SDG frequencies for the different labeling systems. The \code{crosstab\_sdg} function helps reveal patterns of label co-occurrences either across SDGs or systems, which can be controlled using the \code{compare} argument. 

% \usepackage{booktabs}
% \usepackage{graphicx}
\begin{table}[]
\resizebox{\columnwidth}{!}{%
\begin{tabular}{@{}ll@{}}
\toprule
Function Name & Description                                     \\ \midrule
\code{detect\_sdg} & identifies SDGs in text using one of 5 labeling systems (Elsevier, Aurora, SIRIS, OSDG, SDSN). \\
detect\_any &  similar to \code{detect\_sdg} but identifies SDGs in text using user-defined queries. \\
\code{crosstab\_sdg} & crosstab\_sdg takes the output of either detect\_sdg or detect\_any as input and determines correlations between either query systems or SDGs. \\
\code{plot\_sdg} & takes the output of either detect\_sdg or detect\_any and produces adjustable barplots illustrating the hit frequencies produced by the different query systems. \\ \bottomrule
\end{tabular}%
}
\caption{Overview of package functions}
\label{tab:functions_overview}
\end{table}

\section{Demonstrating the functionality of \CRANpkg{text2sdg}}

To showcase the functionalities of the \CRANpkg{text2sdg} package we analyze the publicly available p3 dataset of the Swiss National Science Foundation (SNSF) that lists research projects funded by the SNSF. In addition to demonstrating \CRANpkg{text2sdg}, the case study will permit us to discuss practical issues concerning the labeling of SDGs, including relevant differences between labeling systems. The data and script to reproduce the analyses presented below can be found at \url{https://osf.io/ekzhg/}    

\subsection{Preparing the SNSF projects data}

The SNSF projects data was downloaded from \url{https://data.snf.ch/datasets}. As of March 2022, the p3 database included information on 81,237 research projects. From the data, we removed 54,288  projects where the abstract was absent or not written in English. This left us with a total of 26,949 projects. To ready this data for analysis, we read it using the \code{readr} function of the \CRANpkg{readr} package \citep{readr}, producing a \code{tibble} named \texttt{projects}. A reduced version of this \code{tibble} is included in the \CRANpkg{text2sdg} package. 

\subsection{Using \code{detect\_sdg} to detect SDGs}

To label the abstracts in \code{projects} using \CRANpkg{text2sdg}, we supply the character vector that includes the abstracts to the \code{data} argument of the \code{detect\_sdg} function. Additionally, we make two optional settings. First, we specify that all five systems should be run, rather than the default of only Aurora, Elsevier, and SIRIS, by supplying a character vector of the systems’ names to the \code{systems} argument. Second, we explicitly set the \code{output} argument to \texttt{“features”}, which in contrast to \code{output = “documents”} delivers more detailed information about which keywords triggered the SDG labels.     

\begin{example}
# detect SDGs
> sdgs <- detect_sdg(text = projects, 
+                    systems = c("Aurora", "Elsevier", "SIRIS", "SDSN", "OSDG"), 
+                    output = "features")
    
> head(sdgs)
# A tibble: 6 × 6
  document sdg    system query_id features      hit
     <dbl> <chr>  <chr>     <dbl> <chr>       <dbl>
1        1 SDG-01 SDSN        392 sustainable     4
2        1 SDG-02 SDSN        376 maize           3
3        1 SDG-02 SDSN        629 sustainable     8
4        1 SDG-08 OSDG       3968 work            1
5        1 SDG-08 SDSN        812 work           11
6        1 SDG-09 SDSN        483 research        6
    
\end{example}

The above \code{tibble} produced by \CRANpkg{text2sdg} contains for every combination of document, SDG, system, and query (columns 1 to 4), the query feature (keyword) that triggered the label (columns 5), and a hit index for a given system (column 6). The first row of the \code{tibble} thus shows that the query 392 within SDSN labeled document number 1 with SDG-01, because the document included the feature \textit{sustainable}, and that this was the fourth hit produced by the SDSN system. It is important to note that, in other cases, multiple features of a query might be matched, which will result in multiple rows per combination of document, SDG, system, and query. This can be avoided by setting the \code{output} argument to \texttt{“documents”}, in which case all features' hits of such combinations will be grouped into a single row. 

\subsection{Analyzing the SDG labels}

To visualize the distribution of SDG labels across SDGs and systems in the \texttt{sdgs} \code{tibble}, we apply the \code{plot\_sdg} function. By default, \code{plot\_sdg} shows a barplot of the number of documents labeled by each of the SDGs, with the frequencies associated with the different systems stacked on top of each other. The function counts a maximum of one hit per document-system-SDG combination. Duplicate combinations resulting from hits by multiple queries or keywords in queries will be suppressed by default and the function returns a message reporting the number of cases affected.

\begin{example}

> plot_sdg(sdgs)
138740 duplicate hits removed. Set remove_duplicates = FALSE to retain duplicates.
\end{example}

\begin{figure}[htbp]
  \centering
   \includegraphics[width=1\linewidth]{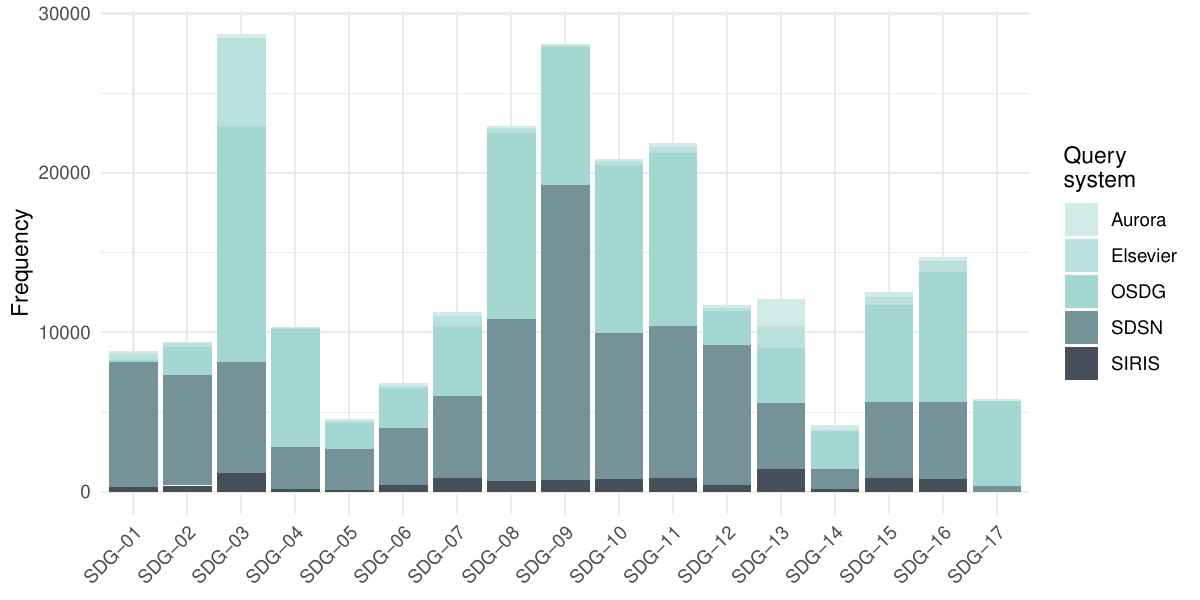}
  \caption{Default plot of distribution of detected SDGs.}
  \label{figure:default_plot}
\end{figure}

The plot produced by \code{plot\_sdg} (Figure ~\ref{figure:default_plot}) shows considerable differences in the frequency of different SDGs, with SDGs 3 (“Good Health and Well-Being”) and 9 (“Industry, Innovation And Infrastructure”) being most frequent and SDGs 5 (“Gender Equality”) and 14 (“Life Below Water”) being least frequent. Furthermore, there are substantial differences in the number of labels produced by different systems, with SDSN and OSDG having produced many more labels than the other three systems. 

\begin{figure}[htbp]
  \centering
   \includegraphics[width=1\linewidth]{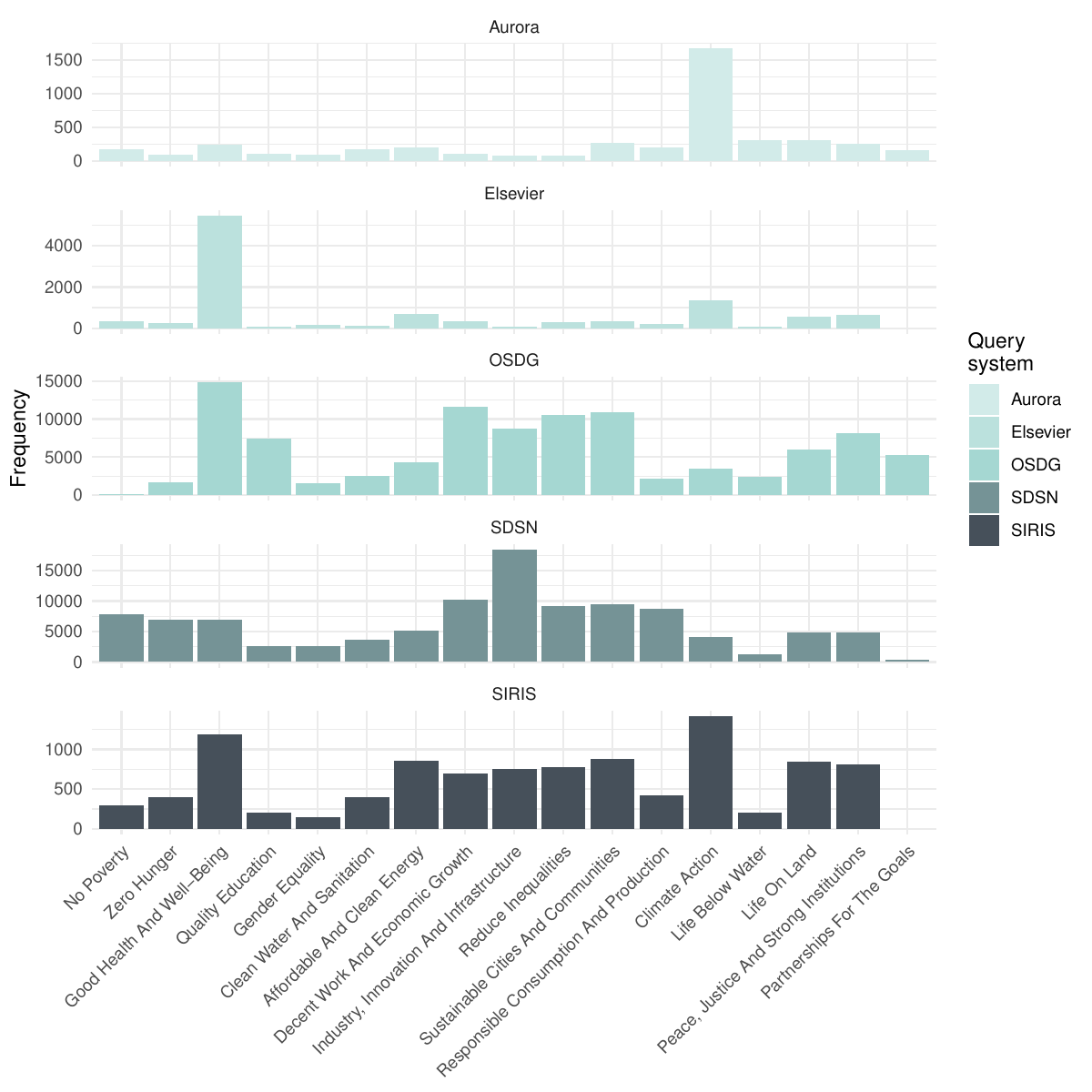}
  \caption{Distribution of detected SDGs facetted by system.}
  \label{figure:default_plot_facetted}
\end{figure}

To customize the visualization of SDG frequencies, the \code{plot\_sdg} function provides several additional arguments. For instance, by setting \code{sdg\_titles} to \texttt{TRUE}, the SDG titles will be added to the annotation of the plot. Other arguments are \code{normalize} to show probabilities instead of frequencies, \code{color} to change the filling of bars, and \code{remove\_duplicates} to eliminate duplicate document-system-SDG combinations. Furthermore, as \code{plot\_sdg} is built on \CRANpkg{ggplot2} \citep{ggplot2}, the function can easily be extended by functions from the \CRANpkg{ggplot2} universe. To illustrate these points, the code below generates a plot (Figure ~\ref{figure:default_plot_facetted}) that includes SDG titles and separates the results of the different SDG systems using facets.

\begin{example}
> plot_sdg(sdgs, 
+          sdg_titles = TRUE) + 
+   ggplot2::facet_wrap(~system, ncol= 1, scales = "free_y")
138740 duplicate hits removed. Set remove_duplicates = FALSE to retain duplicates.
\end{example}

The separation of systems better illustrates the results of systems that produce fewer hits and helps compare the results across systems. This reveals, for instance, that in the Elsevier system SDG 3 (“Good Health and Well-Being”) was most prominent, whereas in the Aurora system this was SDG 13 ("Climate Action”). These results highlight that the different labeling systems do not necessarily agree concerning the assignment of SDGs to documents. 

To quantify the commonalities and differences between labeling systems, \CRANpkg{text2sdg} provides the \code{crosstab\_sdg} function. The function evaluates the level of alignment across either systems (the default) or SDGs by calculating $\phi$ coefficients between the vectors of labels. We supply the \code{hits} argument of the function with the \texttt{sdgs} \code{tibble} containing the labels produced by \code{detect\_sdg}. Note that the function only considers distinct combinations of documents, systems and SDGs, irrespective of whether the \code{detect\_sdg} function was run using \code{output = “documents”} or \code{output = "features”}.    

\begin{example}

> crosstab_sdg(sdgs)

            Aurora  Elsevier      OSDG      SDSN     SIRIS
Aurora   1.0000000 0.3257805 0.1613242 0.1566679 0.3710827
Elsevier 0.3257805 1.0000000 0.2640787 0.2185416 0.3531881
OSDG     0.1613242 0.2640787 1.0000000 0.3705527 0.2242870
SDSN     0.1566679 0.2185416 0.3705527 1.0000000 0.2313963
SIRIS    0.3710827 0.3531881 0.2242870 0.2313963 1.0000000  

\end{example}

The output of \code{crosstab\_sdg()} for the SNSF projects reveals two noteworthy insights. First, the correspondence between the labels of different systems is rather small, as indicated by $\phi$ coefficients that are mostly smaller than .3. Second, there are two groups of systems that are more similar to one another. On the one hand, Elsevier, Aurora, and SIRIS, and, on the other hand, OSDG and SDSN. These groups correspond to differences in query operators, with the former three including AND operators in their queries, whereas the latter two do not.   

It can further be informative to analyze the correlations between SDGs. To do this, we set the \code{compare} in \code{crosstab\_sdg()} to \texttt{"sdgs"}. The output below shows the result for the first six SDGs by setting \code{sdgs = 1:6}. It can be seen that certain pairs of SDGs---in particular, SDG-01 and SDG-02---co-occur more frequently. These results may provides insights into the co-occurrence structure of SDGs in the data at hand. However, these results can also highlight the importance of considering similarities between queries targeting different SDGs.  

\begin{example}

> crosstab_sdg(sdgs, compare = "sdgs", sdgs = 1:6)
           SDG-01     SDG-02     SDG-03     SDG-04     SDG-05     SDG-06
SDG-01 1.00000000 0.50306655 0.05402719 0.07638151 0.14085649 0.17123043
SDG-02 0.50306655 1.00000000 0.10586094 0.07100996 0.09844913 0.18282360
SDG-03 0.05402719 0.10586094 1.00000000 0.21671686 0.12413285 0.06420538
SDG-04 0.07638151 0.07100996 0.21671686 1.00000000 0.12034798 0.08393077
SDG-05 0.14085649 0.09844913 0.12413285 0.12034798 1.00000000 0.04829632
SDG-06 0.17123043 0.18282360 0.06420538 0.08393077 0.04829632 1.00000000 


\end{example}

\section{Practical considerations}

\subsection{Specifying user-defined labeling systems}

 The query systems implemented in \CRANpkg{text2sdg} represent important efforts to systematize the monitoring of SDGs from text. Nevertheless, these efforts are still relatively young and validations of the systems are largely missing, creating a need for continued development. \CRANpkg{text2sdg} supports the further development of new SDG labeling systems by providing the \code{detect\_any} function. In this section, we provide additional detail on using this feature of \CRANpkg{text2sdg}.  

 The \code{detect\_any} function also uses \CRANpkg{corpustools} as the back-end. This implies that new queries must be specified to match the syntax of \CRANpkg{corpustools}. The syntax supports standard Boolean operators (AND, OR, and NOT), wildcard operators, and proximity search. Boolean operators control how different keywords are combined in a query. For instance, the query "marine OR fisheries" matches text that contains either these two words whereas the query "marine AND fisheries" only matches text that contains both these two words. The wildcard operators $?$ and $*$ allow the specification of variable word parts. For instance, the question mark operator $?$ matches one unknown character or no character at all, e.g., "?ish" would match "fish", "dish", or "ish". The asterisk operator $*$, by contrast, matches any number of unknown characters, e.g., "*ish" would match "fish" but also "Swedish". Both wildcards can be used at the start, within or end of a term. Proximity search extends a Boolean AND, by requiring that two keywords have no more than defined distances to one another. For instance, "climate change"~3 specifies matches in which "climate" and "changed" both occur no more than three words apart. A complete description of the \CRANpkg{corpustools} syntax is presented in the \CRANpkg{corpustools} vignette and documentation. 

 To supply a user-defined labeling system to \code{detect\_any}, the queries must be placed in a \code{data.frame} or \code{tibble} that additionally includes a column specifying the labeling system's name and a column of SDG labels corresponding to the queries. 
 
\begin{itemize}
  \item system (character) - name of the labeling systems.
  \item queries (character) - user-defined queries.
  \item sdg (integer) - SDGs labels assigned by queries.
\end{itemize}

 The example below illustrates the application of a user-defined labeling system using \code{detect\_any}. First, a \code{tibble} is defined that includes three rows, one for each of three different queries stored in the \code{query} column. The system is called \texttt{"my\_system"} in the \texttt{system} column and each of the queries is assigned SDG-14 in the \texttt{sdg} column. Note that specification of the labeling system need not be made in R, but can easily be outsourced to a spreadsheet that is then processed into a \code{tibble}. Second, the system is supplied to the \code{system} argument of the \code{detect\_any} function, along with the texts (here, the SNSF abstracts). The output is analogous to the output of the \code{detect\_sdg} function (for brevity, we only show the first three lines of the output).

\begin{example}
> # definition of query set
> my_system <- tibble::tibble(system = "my_system",
+                             query = c("marine AND fisheries", 
+                                        "('marine fisheries') AND sea", 
+                                        "?ish"),
+                             sdg = c(14,14,14))
> detect_any(text = projects, 
+            system = my_system)
# A tibble: 591 × 6
   document sdg    system    query_id features   hit
   <fct>    <chr>  <chr>        <dbl> <chr>    <int>
 1 6        SDG-14 my_system        3 wish       122
 2 134      SDG-14 my_system        3 wish        18
 3 241      SDG-14 my_system        3 fish        59

\end{example}

\subsection{Applying \CRANpkg{text2sdg} to non-English data}
The queries of the labeling systems implemented by \CRANpkg{text2sdg} are in English, implying that texts in other languages must first be translated to English. We assessed whether translation affects the reliability of SDG labels by making use of back translation. To this end, we first translated 1,500 randomly selected SNSF project abstracts from English to German and from German to English and then compared the labels of the original English and back-translated English abstracts. We carried out the translation using the DeepL translation engine (\href{https://www.deepl.com/translator}{www.deepl.com/translator}). 

Table \ref{tab:my-table_corr} shows the results of this analysis. Overall, the correlations as measured by the $phi$-coefficient are very high. The systems showed correlations above or equal to $0.88$, with Elsevier showing the highest value of $.93$. Considering that our analysis involves not only one, but two translation steps---from German to English and back---these results suggest that \CRANpkg{text2sdg} can be applied to non-English text with very high accuracy. One should note, however, that the quality of translation may vary between languages and translations engines. 

% \usepackage{booktabs}
% \usepackage{graphicx}
\begin{table}[h]
\centering
\begin{tabular}{@{}lllll@{}}
\toprule
Aurora & Elsevier & SIRIS & SDSN & OSDG \\ \midrule
.91 & .93 & .88 & .91 & .91 \\
\bottomrule
\end{tabular}%
\caption{$phi$-coefficient between the labels for the original English text and the labels for the back-translated (English-German-English) English text}
\label{tab:my-table_corr}
\end{table}

 \subsection{Estimating the runtime of \CRANpkg{text2sdg}}

 The analysis of text data can be computationally intense. To provide some guidance on the expected runtime of \CRANpkg{text2sdg} for data with different numbers of documents and different document lengths, we carried out several experiments. For this purpose, we first simulated documents by concatenating 10, 100, 1,000, or 10,000 words drawn randomly according to word frequencies in Wikipedia and combined 1, 10, 100, or 1,000 thus-generated documents into simulated data sets. Then we evaluated the runtime of \CRANpkg{text2sdg} separately by system for the simulated data sets. 
 
\begin{figure}[htbp]
  \centering
   \includegraphics[width=1\linewidth]{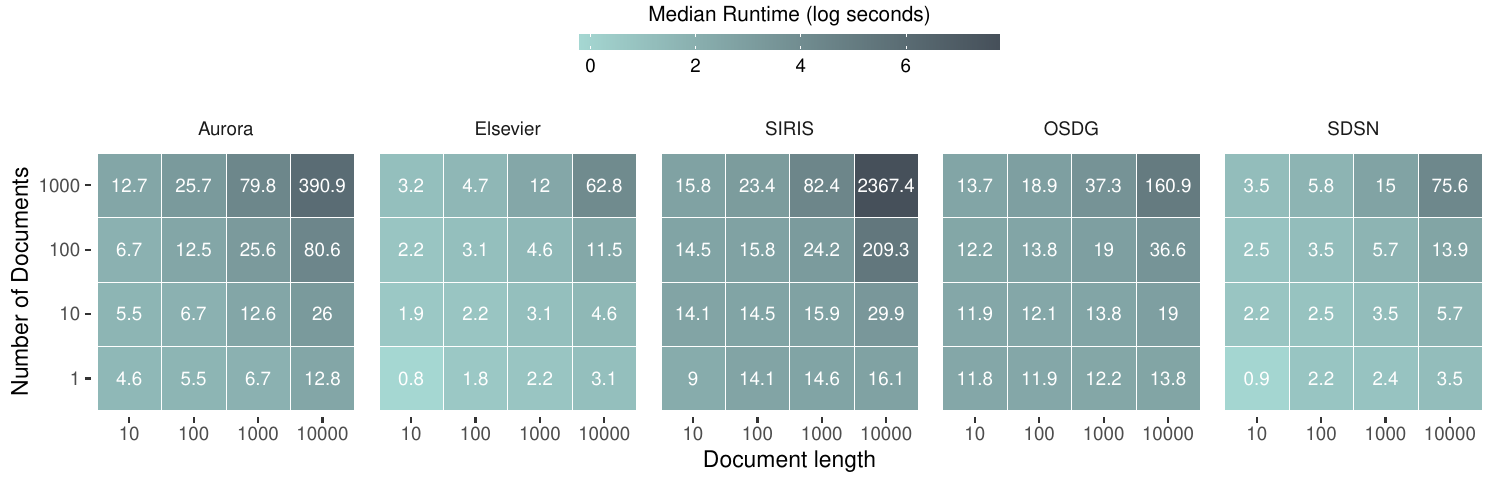}
  \caption{Runtime per seconds averaged across 7,000 runs for each cell. Color reflects log runtime.}
  \label{figure:benchmark_plot}
\end{figure}

 Figure~\ref{figure:benchmark_plot} shows the average runtime in seconds across 7,000 repetitions of each combination of document length and number of documents for each of the labeling systems. The results highlight noteworthy points. First, runtime is primarily a function of the number of words, irrespective of how words are distributed across documents. Second, the runtime per words decreases as the number of words increases, which is due to a constant overhead associated with optimizing the labeling systems' queries. Third, there are considerable differences in the runtime between systems, which is, in part, due to the functions' overhead and, in part, due to differences in number and complexity of queries. The fastest system is Elsevier, processing 10 million words in roughly one minute; the slowest system is SIRIS, processing 10 million words in about 40 minutes. 
 Overall, these experiments highlight that \CRANpkg{text2sdg} can efficiently process large amounts of text, but also that some care should be exercised when dealing with extremely large or many texts. In such cases, it may be advisable to rely on more efficient labeling systems, such as Elsevier or SDSN.   

\section{Conclusion}

 In this article, we introduced a new R package, \CRANpkg{text2sdg}, designed to help identify SDGs from text. The package promises to help detect SDGs in any kind of text sources using different existing or custom-made labeling systems. Our case study and additional analyses suggest that the approach can handle both sources in English as well as translations, allows user-friendly use of novel queries, and provides reasonably efficient performance for analysing large corpora. 

\bibliography{text2sdg}

\begin{thebibliography}{10}
\providecommand{\natexlab}[1]{#1}
\providecommand{\url}[1]{\texttt{#1}}
\expandafter\ifx\csname urlstyle\endcsname\relax
  \providecommand{\doi}[1]{doi: #1}\else
  \providecommand{\doi}{doi: \begingroup \urlstyle{rm}\Url}\fi

\bibitem[Bautista(2019)]{Bautista2019}
N.~Bautista.
\newblock Sdg ontology.
\newblock 2019.
\newblock URL \url{https://doi.org/10.6084/m9.figshare.11106113.v1}.

\bibitem[Duran-Silva et~al.(2019)Duran-Silva, Fuster, Massucci, and
  Quinquillà]{duran_silva_nicolau_2019_3567769}
N.~Duran-Silva, E.~Fuster, F.~A. Massucci, and A.~Quinquillà.
\newblock {A controlled vocabulary defining the semantic perimeter of
  Sustainable Development Goals}, Dec. 2019.
\newblock URL \url{https://doi.org/10.5281/zenodo.3567769}.

\bibitem[Jayabalasingham et~al.(2019)Jayabalasingham, Boverhof, Agnew, and
  Klein]{jayabalasingham2019identifying}
B.~Jayabalasingham, R.~Boverhof, K.~Agnew, and L.~Klein.
\newblock Identifying research supporting the united nations sustainable
  development goals.
\newblock \emph{Mendeley Data}, 1, 2019.
\newblock URL \url{https://doi.org/10.17632/87txkw7khs.1}.

\bibitem[{Sustainable Development Solutions Network (SDSN)}(2021)]{sdsn}
{Sustainable Development Solutions Network (SDSN)}.
\newblock Compiled list of sdg keywords, 2021.
\newblock URL
  \url{https://ap-unsdsn.org/regional-initiatives/universities-sdgs/}.

\bibitem[UN(2022)]{SGD_report2022}
UN.
\newblock \emph{{The Sustainable Development Goals Report 2022}}.
\newblock United Nations, 2022.

\bibitem[Vanderfeesten et~al.(2020{\natexlab{a}})Vanderfeesten, Otten, and
  Spielberg]{vanderfeesten_maurice_2020_3817445}
M.~Vanderfeesten, R.~Otten, and E.~Spielberg.
\newblock {Search Queries for "Mapping Research Output to the Sustainable
  Development Goals (SDGs)" v5.0}, July 2020{\natexlab{a}}.
\newblock URL \url{https://doi.org/10.5281/zenodo.3817445}.

\bibitem[Vanderfeesten et~al.(2020{\natexlab{b}})Vanderfeesten, Spielberg, and
  Gunes]{vanderfeesten_maurice_2020_3813230}
M.~Vanderfeesten, E.~Spielberg, and Y.~Gunes.
\newblock {Survey data of "Mapping Research Output to the Sustainable
  Development Goals (SDGs)"}, May 2020{\natexlab{b}}.
\newblock URL \url{https://doi.org/10.5281/zenodo.3813230}.

\bibitem[Welbers and {van Atteveldt}(2021)]{corpustools}
K.~Welbers and W.~{van Atteveldt}.
\newblock \emph{corpustools: Managing, Querying and Analyzing Tokenized Text},
  2021.
\newblock URL \url{https://CRAN.R-project.org/package=corpustools}.
\newblock R package version 0.4.8.

\bibitem[Wickham(2016)]{ggplot2}
H.~Wickham.
\newblock \emph{ggplot2: Elegant Graphics for Data Analysis}.
\newblock Springer-Verlag New York, 2016.
\newblock ISBN 978-3-319-24277-4.
\newblock URL \url{https://ggplot2.tidyverse.org}.

\bibitem[Wickham et~al.(2021)Wickham, Hester, and Bryan]{readr}
H.~Wickham, J.~Hester, and J.~Bryan.
\newblock \emph{readr: Read Rectangular Text Data}, 2021.
\newblock URL \url{https://CRAN.R-project.org/package=readr}.
\newblock R package version 2.1.1.

\end{thebibliography}

\address{Dominik S. Meier\\
  University of Basel\\
  Steinengraben 22 4051 Basel\\
  Switzerland\\
  (ORCID: 0000-0002-3999-1388)\\
  \email{dominik.meier@unibas.ch}}

\address{Rui Mata\\
  University of Basel\\
  Missionsstrasse 60-62 4055 Basel\\
  Switzerland\\
  (ORCID: 0000-0002-1679-906X)\\
  \email{rui.mata@unibas.ch}}
  
\address{Dirk U. Wulff\\
  University of Basel\\
  Missionsstrasse 60-62 4055 Basel\\
  Switzerland\\
  (ORCID: 0000-0002-4008-8022)\\
  \email{dirk.wulff@unibas.ch}}

\end{article}

\end{document}